\documentclass[conference]{IEEEtran}
\IEEEoverridecommandlockouts
\usepackage{amsmath,amssymb,amsfonts}
\usepackage{algorithmic}
\usepackage{graphicx}
\usepackage{textcomp}
\usepackage{multirow}
\usepackage[table,xcdraw]{xcolor}
\usepackage[
backend=biber,
style=numeric,
sorting=none
]{biblatex}
\usepackage{xcolor}
\def\BibTeX{{\rm B\kern-.05em{\sc i\kern-.025em b}\kern-.08em
    T\kern-.1667em\lower.7ex\hbox{E}\kern-.125emX}}
\DeclareUnicodeCharacter{0308}{\"{u}}

\begin{document}

\title{Emotion Classification in Short English Texts using Deep Learning Techniques\\
{\footnotesize}
\thanks{Identify applicable funding agency here. If none, delete this.}
}

\author{\IEEEauthorblockN{1\textsuperscript{st} Siddhanth Bhat}
\IEEEauthorblockA{\textit{School of Computer Science and Engineering} \\
\textit{Manipal University Jaipur}\\
Jaipur, India \\
0009-0005-6350-6601}
}
\maketitle

\begin{abstract}
Detecting emotions in limited text datasets from under-resourced languages presents a formidable obstacle, demanding specialized frameworks and computational strategies. This study conducts a thorough examination of deep learning techniques for discerning emotions in short English texts.Deep learning approaches employ transfer learning and word embedding, notably BERT, to attain superior accuracy. To evaluate these methods, we introduce the "SmallEnglishEmotions" dataset, comprising 6372 varied short English texts annotated with five primary emotion categories. Our experiments reveal that transfer learning and BERT-based text embedding outperform alternative methods in accurately categorizing the text in the dataset.
\end{abstract}


\begin{figure*}[htbp!]
\centerline{\includegraphics[width =7 in]{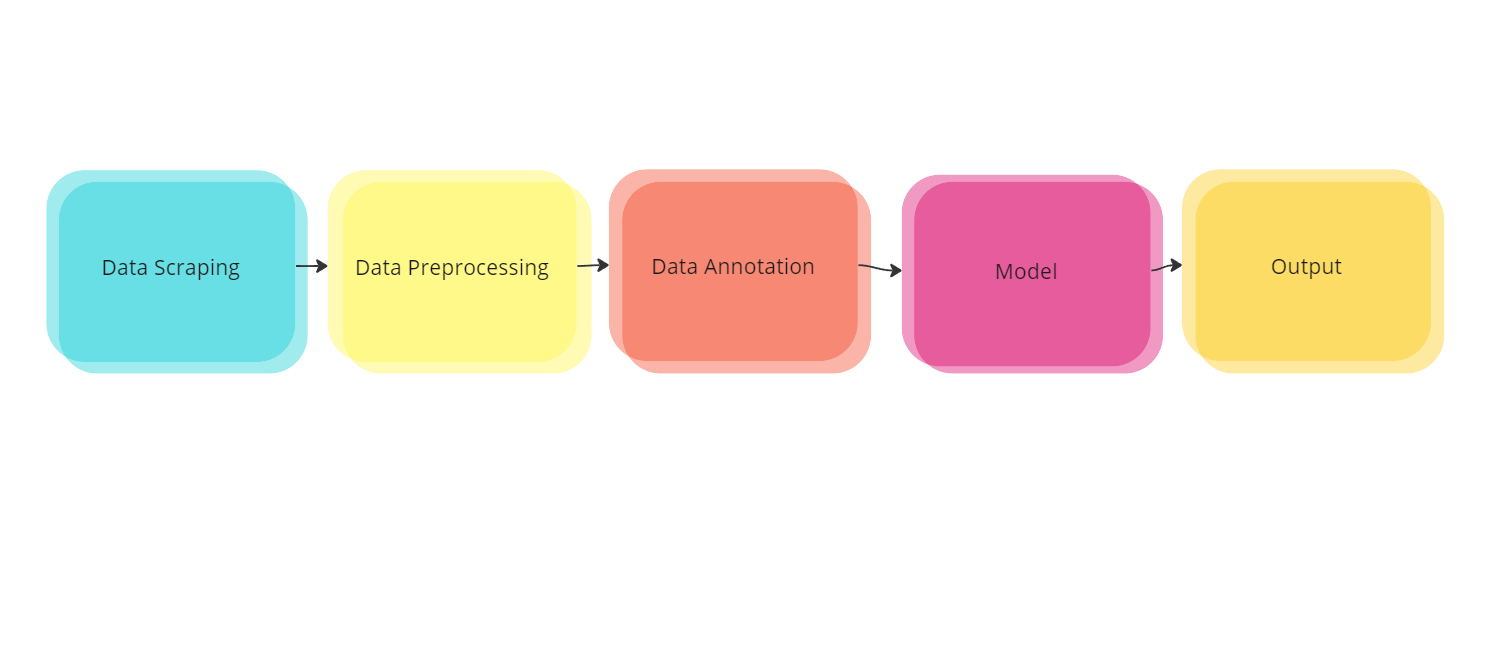}}
\caption{Proposed Methodology Architecture}
\label{fig}
\end{figure*}

\section{Introduction}
In today's digital age, the pervasive influence of online social media platforms has revolutionized the way individuals communicate, share ideas, and express their innermost thoughts and emotions \cite{peng2022survey}. The virtual landscape is teeming with a myriad of textual content ranging from personal anecdotes to political discourse, creating a rich tapestry of human expression to be deciphered and understood. Harnessing the power of deep learning techniques for sentiment analysis and emotion recognition of textual data has emerged as a critical tool in decoding the underlying sentiments and emotional nuances embedded within.

This analytical approach serves as a compass for navigating the sea of human emotions, offering valuable insights into the collective psyche of individuals and communities alike \cite{peng2022survey,acheampong2020text,kratzwald2018deep,lakizadeh2022text}. From assessing the ripple effects of pivotal decisions made by industry magnates and policymakers to unraveling the customer satisfaction levels in the realm of online commerce, the applications of sentiment analysis and emotion recognition are as diverse as they are important. Moreover, in the realm of social and political discourse, these tools serve as invaluable instruments for dissecting the sentiments and reactions of online communities in response to pivotal events and societal upheavals.

Beyond the realm of public discourse, sentiment analysis and emotion recognition also play a pivotal role in the realm of mental health, offering a lifeline to individuals grappling with psychological disorders \cite{balakrishnan2019string,ghosh2023love,nasiri2021denova}. By deciphering the emotional cues embedded within textual data, these systems provide clinicians with valuable insights into the emotional well-being of their patients, aiding in the formulation of personalized treatment plans and interventions.

Furthermore, in the arena of governance and policy-making, sentiment analysis serves as a powerful tool for gauging public sentiment towards political figures, administrations, and policy initiatives \cite{balakrishnan2019string,ghosh2023love,nasiri2021denova}. By analyzing the sentiments expressed across various online platforms, policymakers gain valuable insights into the public's perceptions and attitudes, enabling them to tailor their policies and initiatives to better align with the needs and aspirations of their constituents.

\subsection{Related Work}
Recent research has highlighted the effectiveness of deep learning methodologies, such as transfer learning and word embedding, in recognizing emotions in languages with limited resources. Transfer learning, particularly in the realm of text embedding, has been leveraged to customize existing models for emotion detection tasks. Various iterations of BERT and similar models have been employed for text embedding purposes. Moreover, transfer learning techniques have been applied to enhance the comprehension of emotional word semantics within deep learning frameworks. Researchers also utilize approaches like combining models from diverse linguistic backgrounds and employing data augmentation techniques to enhance the accuracy of emotion recognition systems. In the domain of English text emotion recognition, early studies primarily relied on shallow learning approaches but gradually transitioned to more sophisticated methods like GRU and BERT-based embedding models such as XLM-R \cite{kumar2021sentiment}. Prior investigations often failed to consider the impact of text length on classification complexity, despite its acknowledged significance. Hence, further exploration is warranted to meet the challenges associated with detecting emotions in concise English texts.

\subsection{Motivation and Requirement}
The goal of this research is to propose a novel training model to the BERT language model with standard statistical text models of character, word, and structure levels for the single-level topical classification for the gathered text dataset. Standard models are based on the letters and punctuation mark occurrences, word and parts-of-speech n-grams.
\section{Proposed Methodology}

\subsection{Dataset}
The study introduces SmallEnglishEmotions, a novel dataset comprising concise English texts labeled with five distinct emotional categories. The choice of short texts for emotion detection is rooted in the scientific rationale that single-emotion presence in each short text simplifies annotation for evaluators, facilitating more precise assessment by deep learning algorithms. Additionally, reducing the number of emotions per text enhances automatic emotion recognition confidence, yielding more robust outcomes. Furthermore, the brevity of short texts poses challenges for both human readers and deep learning models, urging researchers to devise improved classification methods. Thus, this benchmark dataset presents a more demanding task, encouraging researchers to develop and evaluate emotion classification techniques with heightened assurance. This segment provides a detailed description of the SmallEnglishEmotions dataset along with its principal attributes.

\subsection{Dataset Collection}
The dataset employed in this study, known as SmallEnglishEmotions, comprises of 6372 short English texts sourced from Twitter. Annotation of the dataset adheres to Rachael Jack's emotional model [31], categorizing texts into five emotional classes: happiness, sadness, anger, fear, and others. Unlike many publicly available datasets, SmallEnglishEmotions focuses specifically on shorter texts, with an average length of 50 words.
\\
Data collection involved utilizing the Snscrape library in Python to gather tweets from Twitter selectively. Tweets were sourced around specific topics such as sports, politics, trending news, and the USA, among others. Each topic was associated with a specific query containing relevant keywords. Data collection spanned the years 2020 to 2023, with texts initially not restricted by length. Subsequently, texts ranging from 20 to 70 words were selected as short texts for inclusion in the dataset. Geographical restrictions were not imposed, but only English tweets were considered during the search process.
.
\subsection{Dataset Labelling}
Each entry within the dataset is associated with a single emotional label. Human annotators, numbering five in total, were tasked with assigning emotional labels to each sample. This annotation process was facilitated through a web-based application developed using React and a CSV database. Upon completion of the annotation task, the collected data were stored in the database and subsequently retrieved into a pandas dataframe for analysis. Each annotator was assigned a unique identifier (ID) to track their contributions. Using this ID and the provided user interface, annotators navigated through the dataset, assigning emotions to each sample by selecting from a range of predefined options. Annotators were instructed to exercise caution, avoiding the assignment of multiple labels or labeling ambiguous instances by utilizing the provided "ignore" button.

\subsection{Dataset Analysis}
The SmallEnglishEmotions dataset contains 6372 instances in five emotion classes happiness, sadness, fear, anger, and other. The number of instances of each class in the SmallEnglishEmotions dataset is shown in table 1

\begin{table}[!hbtp]
\caption{}
\label{tab:my-table}
\centering
\begin{tabular}{|l|l|}
\hline
\textbf{Emotion} & \textbf{Number of Samples} \\ \hline
Happiness        & 1125                       \\ \hline
Sadness          & 1463                       \\ \hline
Fear             & 1256                       \\ \hline
Anger            & 1125                       \\ \hline
Other            & 1403                       \\ \hline
\end{tabular}
\end{table}

\section{Experimental Analysis}
In this segment, we will delve into the outcomes stemming from various experiments conducted to assess the efficacy of deep learning models on the SmallEnglishEmotions dataset. Moreover, we'll compare the classification proficiency attained on English texts within the SmallEnglishEmotions dataset with that achieved on lengthier Persian texts from a standard twitter dataset. Additionally, aside from evaluating classification performance, we'll also present insights into the training duration for each approach.

\subsection{Hyperparameter Tuning and Environment}
All methodologies proposed in this investigation were implemented utilizing the Python programming language, employing the Scikit Learn and Keras libraries. The experiments were conducted within a Google Colab cloud environment, equipped with 12.7 GB RAM, a T4 GPU, and 15.0 GB GPU RAM. For training the model, the Adam optimizer was employed alongside a batch size of 32. In order to mitigate the risks of underfitting, overfitting, and prolonged training durations, early stopping mechanisms were integrated during the training phase of the model. Specifically, a maximum of 100 epochs with a patience of 10 were specified.The input layer's maximum sentence length was set to 100 words for the SmallEnglishEmotions dataset.

\section{Results and Optimisations}
To assess the efficacy of the proposed techniques in detecting emotions within short English texts, each approach underwent evaluation on the SmallEnglishEmotions dataset. Furthermore, for comparative analysis between emotion classification in short and longer English texts, these methods were also assessed using a standard Twitter dataset. A comparison between the performance of each model on the SmallEnglishEmotions dataset and its performance on the standard dataset was subsequently conducted.

In our experimentation, the proposed machine learning models underwent training and evaluation cycles ten times using the SmallEnglishEmotions dataset. This same experiment was replicated with the standard dataset. The aggregated results are presented in Table 2. Notably, the deep learning-based methods consistently outperformed other learning methods in terms of both Accuracy and Macro-F1 metrics. Particularly, the transfer learning approach utilizing  pretrained distilBERT \cite{sanh2019distilbert} embeddings exhibited the highest accuracy across both datasets, as measured by both criteria. The success of the distilBERT method can be attributed to its ability to consider contextual information by examining surrounding words, thus enhancing the model's accuracy.
\begin{table}[]
\caption{}
\label{tab:my-table}
\resizebox{\columnwidth}{!}{%
\begin{tabular}{|l|l|l|l|l|}
\hline
              & \multicolumn{2}{c|}{Macro-F1}  & \multicolumn{2}{c|}{Accuracy}  \\ \hline
Model Name    & Stnadard &SmallEnglishEmotions & Standard & SmallEnglishEmotions \\ \hline
Reinforcement     & 35\%       & 57\%               & 33\%       & 62\%               \\ \hline
SVM   & 43\%       & 58\%               & 54\%       & 65\%               \\ \hline
BERT & 45\%       & 63\%               & 46\%       & 67\%               \\ \hline
Pretrained DistilBERT & 67\%       & 73\%               & 71\%       & 77\%          \\    \hline
\end{tabular}%
}
\end{table}
\section{Conclusion and Future Work}
The research article investigated emotion classification in short English texts through a variety of statistical and deep learning methodologies. A meticulously curated dataset of short English texts, annotated with five distinct emotion categories, was employed to support the study. The evaluation of both shallow and deep learning models on the SmallEnglishEmotions dataset underscored the superior accuracy of deep learning architectures, particularly those harnessing semantic embedding and transfer learning techniques, in adeptly classifying emotions. Notably, the utilization of the pretrained distilBERT model \cite{sanh2019distilbert} in the deep learning framework emerged as the most effective strategy for achieving remarkable performance in English emotion recognition within short texts.

Moreover, a comparative analysis of model performance on the SmallEnglishEmotions and a standard English dataset revealed a heightened accuracy in classifying shorter texts compared to longer ones. This phenomenon is attributed to the reduced ambiguity inherent in labeling shorter text samples. Future research avenues may focus on exploring innovative methodologies to further enhance the understanding of semantic intricacies embedded within short texts.

\printbibliography
\end{document}